# Contextual Normalization Applied to Aircraft Gas Turbine Engine Diagnosis

## NRC #35028

Peter Turney and Michael Halasz
Knowledge Systems Laboratory
Institute for Information Technology
National Research Council Canada
Ottawa, Ontario, Canada, K1A 0R6
turney@ai.iit.nrc.ca, halasz@ai.iit.nrc.ca

## Abstract

Diagnosing faults in aircraft gas turbine engines is a complex problem. It involves several tasks, including rapid and accurate interpretation of patterns in engine sensor data. We have investigated *contextual normalization* for the development of a software tool to help engine repair technicians with interpretation of sensor data. Contextual normalization is a new strategy for employing machine learning. It handles variation in data that is due to contextual factors, rather than the health of the engine. It does this by normalizing the data in a context-sensitive manner. This learning strategy was developed and tested using 242 observations of an aircraft gas turbine engine in a test cell, where each observation consists of roughly 12,000 numbers, gathered over a 12 second interval. There were eight classes of observations: seven deliberately implanted classes of faults and a healthy class. We compared two approaches to implementing our learning strategy: linear regression and instance-based learning. We have three main results. (1) For the given problem, instance-based learning works better than linear regression. (2) For this problem, contextual normalization works better than other common forms of normalization. (3) The algorithms described here can be the basis for a useful software tool for assisting technicians with the interpretation of sensor data.

# 1  Introduction

This paper discusses the investigation of a learning strategy that we call *contextual normalization*. Contextual normalization is a strategy for using machine learning algorithms for supervised learning from examples. This learning strategy was developed and tested using observations of an aircraft gas turbine engine in a test cell.

When repairing an aircraft gas turbine engine, it is a standard procedure to remove the engine from the plane, mount it in a test cell, then monitor the engine as it runs. This procedure can generate massive amounts of data, very few of which are currently used by the repair technician. We plan to make these data more useful.

The end product of this line of research will be software that maintains a library of examples of different faults. It will take data from an engine, compare them with the library, then inform the technician of the state of the engine. The engine might be healthy, it might have a fault that is in the library, or it might have a new fault. If the data are sufficiently interesting, they may be added to the library. This software could advise a technician directly, or it could be used in conjunction with a knowledge-based system, such as the system described in [1].

In Section 2, we describe our data. We have collected a library of data consisting of 242 observations of an engine in a test cell. These 242 observations fall into eight distinct classes: seven different classes of deliberately implanted faults and a class with no faults (healthy).

In Section 3, we present our strategy for analyzing the data. We developed a three phase procedure to analyze the data. Phase 0 extracts features from the data. Phase 1 normalizes the features. Phase 2 classifies the normalized features. Phase 0 is a hand-tuned procedure that does not currently involve learning. Phases 1 and 2 involve supervised learning from examples.

The novel aspect of this three phase procedure is Phase 1. This phase performs *contextual normalization*. The features from Phase 0 are normalized in Phase 1 in a way that is sensitive to the context of the features. For an aircraft gas turbine engine, the context is the operating regime of the engine and the ambient conditions of the external environment. The purpose of contextual normalization is to handle variation in data that is due to contextual factors, rather than the health of the engine.

The three phase learning strategy of Section 3 can be implemented using many different learning algorithms. Section 4 discusses the implementation of the analysis strategy using instance-based learning (IBL) and Section 5 discusses the implementation using multivariate linear regression (MLR). We call the first implementation CNIBL (Contextual Normalization with Instance-Based Learning) and the second implementation CNMLR (Contextual Normalization with Multivariate Linear Regression).

Instance-based learning is described in [2, 3]. It is closely related to the nearest neighbor pattern recognition paradigm [4]. Predictions are made by matching new data to stored data (called *instances*), using a measure of similarity to find the best matches. We use IBL to predict both real numbers [2] and class membership [3].

Multivariate linear regression is the most popular traditional statistical technique for analyzing data [5]. MLR models data with a system of linear equations. We use MLR to predict both real numbers and class membership. When MLR is used to predict class membership, it is known as discriminant analysis.

Section 6 presents the results of a series of experiments that we performed. We have three main results:

1. For the given problem, instance-based learning works better than linear regression.

2. For the given problem, contextual normalization works better than several other commonly used forms of normalization.

3. Instance-based learning with contextual normalization can be the basis for a useful software tool.

Section 7 examines related work. Finally, in Sections 8 and 9, we discuss future work and our conclusions. Our results support the value of our learning strategy for aircraft gas turbine engine diagnosis.

## 2  Aircraft Gas Turbine Engine Data

We have collected 242 observations of a single aircraft gas turbine engine. The chosen engine (a General Electric J85-CAN-15) is a typical older-generation engine. It is used on the CF5 aircraft of the Canadian Department of National Defence. The observations were collected under a variety of ambient (weather dependent) conditions. With each observation, the engine was either healthy or had one of seven different deliberately implanted faults. Five of the seven faults had varying degrees of severity; they were not merely present or absent. Table 1 is a list of the eight classes of observations. The meaning of the various classes is not relevant here, although it may be worth noting that class 8 is healthy.

Place Table 1 here.

The engine was mounted in a test cell and several sensors were installed. We focussed on faults that appear during acceleration, since this is a common and challenging class of faults for technicians to diagnose. We examined two types of accelerations, rapid acceleration from idle to military power setting and rapid acceleration from idle to maximum afterburner power setting.

Table 2 shows the variables that were recorded for each observation of the aircraft gas turbine engine. Variables 18 to 22 are constant for a single observation. These last five variables record the ambient conditions of the observation. Statistical tests persuaded us to focus on 15 of our 22 variables. Feature extraction (see Section 3.1) uses the 10 variables that are marked with * in Table 2. Feature normalization (see Section 3.2) uses these 10 plus the 5 variables that are marked with **.

Place Table 2 here.

A single observation consists of roughly 12,000 numbers: variables 1 to 17 sampled at 60 Hertz for 12 seconds and variables 18 to 22 sampled once for each observation. During an observation, the throttle (PLA) is rapidly moved (within one second) from idle position to either military or maximum afterburner position.

## 3  Strategy for Data Analysis

This section presents the strategy that was used to analyze the data. The actual implementation of the strategy is discussed in Sections 4 and 5.

The data were analyzed in three phases. In Phase 0, features were extracted from the data. In Phase 1, the extracted features were normalized. In Phase 2, the normalized features were assigned to one of the eight classes. Figure 1 summarizes the strategy for data analysis.

Place Figure 1 here.

## 3.1  Phase 0: Feature Extraction

Of the 22 variables, 9 were chosen as having special importance for diagnosis, based on statistical tests and consultation with engine experts. These variables were THRUST, WFM1, WFT, N1, P3S, P5T, T5, NPI, and IGV. We examined plots of these variables, where the y-axis was one of the variables and the x-axis was TIME. Working with aircraft gas turbine engine experts, we identified features that seemed to have diagnostic significance. These features significantly extend the set of features that are given by the engine manufacturer. Figure 2 shows the features for an acceleration of a healthy engine from idle to maximum afterburner power setting. The features are indicated by boxes in the plots. A feature is defined as a certain shape of curve, such as "the peak in T5". A feature is not defined by a certain *x* or *y* value. The *x* and *y* values of a feature tend to vary from one observation to another. This variation has diagnostic significance.

Place Figure 2 here.

A procedure was written to extract the features from the data. This procedure does not involve learning. For each feature, there is a section of the program that is devoted to finding that feature, by examining the slope of the curve and checking the slope against hand-tuned thresholds. These feature detection routines sometimes do not find a feature and sometimes make mistakes, so the output of Phase 0 includes missing and erroneous values.

For each feature that is not missing, two numbers are generated, the positions of the feature on the x-axis and the y-axis. Different sets of features are extracted for the two types of accelerations. The output of Phase 0 is a vector of length 84 or 126. The elements of the vector are the *x* and *y* values of the 42 (for idle to military accelerations) or 63 (for idle to maximum afterburner acceler-ations) features.

The main function of Phase 0 is to compress the data from a cumbersome set of about 12,000 numbers to a more manageable set of about 100 numbers. Although information is lost, the compression is designed to preserve the information that is required for successful diagnosis.

## 3.2 Phase 1: Contextual Normalization of Features

Let $\vec{v}$ be a vector of features, $(v_1, \ldots, v_n)$. In Phase 1, we wish to transform $\vec{v}$ to a vector $\vec{\eta}$ of normalized features, $(\eta_1, \ldots, \eta_n)$. We use the following formula to normalize the features:

$$\eta_i = \frac{v_i - \mu_i(\vec{c})}{\sigma_i(\vec{c})} \qquad \text{(EQ 1)}$$

The context in which $\vec{v}$ was observed is given by the context vector $\vec{c}$. For our application, the context is given by the ambient conditions, T1, TEX, TDEW, BARO, and HUMID (respectively, the inlet air temperature, the outside air temperature, the dew point temperature, the outside air pressure, and the relative humidity). The expected value of $v_i$ as a function of the context is $\mu_i(\vec{c})$. The expected variation of $v_i$ as a function of the context is $\sigma_i(\vec{c})$. Equation 1 is similar to normalizing $v_i$ by subtracting the average and dividing by the standard deviation, except that $\mu_i(\vec{c})$ (analogous to the average) and $\sigma_i(\vec{c})$ (analogous to the standard deviation) are functions of the context $\vec{c}$. Aircraft gas turbine engines are very sensitive to ambient conditions, so it is natural to calculate the expected value and expected variation of $v_i$ as functions of $\vec{c}$. (A list of symbols is included as an appendix.)

The expected value $\mu_i(\vec{c})$ and the expected variation $\sigma_i(\vec{c})$ of a feature $v_i$ are calculated from a set of 16 healthy observations, which we call our *baseline* observations. Thus $\mu_i(\vec{c})$ and $\sigma_i(\vec{c})$ are the expected value and variation of $v_i$ in the context $\vec{c}$, *assuming that the engine is healthy*. The baseline observations were chosen to span a wide range of ambient conditions. We have one set of 16 baseline observations for idle to military accelerations and a second set of 16 baseline observations for idle to maximum afterburner accelerations.

Note that the normalization of Equation 1 is centered on zero and can potentially range from minus to plus infinity. The output of Phase 1 is a normalized feature vector $\vec{\eta}$. The length of the output vector is the same as the length of the input feature vector $\vec{v}$.

Phase 1 has three characteristics:

1. The normalized features all have the same scale, so we can directly compare such disparate things as THRUST and T5. IBL works best with numbers that have been normalized to the same scale [3].

2. Equation 1 tends to weight features according to their relevance for diagnosis. Features that do not fit with expectations, given the baselines, are normalized to values that are relatively far from zero. That is, a surprising feature will get a high absolute value. A feature that is irrelevant will tend to have a high expected variation in the baselines, so it will tend to be normalized to a value near zero.

3. Equation 1 compensates for performance variations that are due to variations in the ambient conditions. This is why the baseline observations were chosen to span a range of ambient conditions.

As we shall see in Section 6.3, Phase 1 is superior to every alternative normalization procedure that we have examined. None of the alternatives have all three of these characteristics.

### 3.3  Phase 2: Diagnosis

The function of Phase 2 is to make a diagnosis. The output currently takes the form of a single predicted class. The 242 observations are split into training and testing sets. Normalized feature vectors in the testing set, with unknown status, are classified by examining vectors in the training set, with known status. In other words, the normalized feature vector $\vec{\eta}$ for the current observation is compared with a library of normalized feature vectors from past observations.

## 4  Instance-Based Learning

Instance-based learning [2, 3] was used to implement Phases 1 and 2 of the strategy described above. Phase 1 requires prediction of real numbers and Phase 2 requires prediction of class membership. Both phases involve supervised learning from examples. Instance-based learning is a natural candidate, since it is a simple technique that meets these requirements. We call the IBL implementation of our strategy CNIBL.

### 4.1  Phase 1: Feature Normalization with IBL

Suppose we have a feature vector $\vec{v}$ that we wish to normalize, using Equation 1. We need to find the expected value $\mu_i(\vec{c})$ and expected variation $\sigma_i(\vec{c})$ of each element $v_i$ of $\vec{v}$. The 16 baseline observations are processed through Phase 0, so we have 16 feature vectors. These are our instances. We find the expected value $\mu_i(\vec{c})$ of $v_i$ by looking for the $k_1$ instances with ambient conditions that most closely match the ambient conditions $\vec{c}$ of $\vec{v}$. We found that the optimal value of $k_1$ is 2. (We discuss experiments with parameter settings in Section 6.2.)

In order to measure the similarity between ambient conditions, we first normalize the ambient conditions. Each of the five ambient condition variables is normalized by its minimum and maximum value in the 16 baseline observations. The minimum is normalized to zero and the maximum is normalized to one. The ambient conditions $\vec{c}$ for $\vec{v}$ are also normalized by the minimum and maximum in the baseline observations. With some experimentation, we chose the following measure of similarity between vectors $\vec{x}$ and $\vec{y}$

$$\sum_i (1 - \text{abs}(x_i - y_i)) \tag{EQ 2}$$

where $\vec{x}$ and $\vec{y}$ are arbitrary normalized vectors and abs() is the absolute value function. We apply this measure of similarity to the normalized ambient conditions for $\vec{v}$ and for each baseline, in order to find the $k_1$ baselines that are most similar to $\vec{v}$.

Let $K_1$ be the set of the $k_1$ baselines that are most similar to $\vec{v}$. For each element $v_i$ of $\vec{v}$, the expected value $\mu_i(\vec{c})$ of the element $v_i$ is the weighted average of the value of the corresponding element in each of the members of $K_1$. The weights are linearly proportional to the degrees of similarity between the ambient conditions $\vec{c}$ for $\vec{v}$ and the ambient conditions for the members of $K_1$.

We find the expected variation $\sigma_i(\vec{c})$ of $v_i$ by looking for the $k_2$ instances whose ambient conditions are the *next* most similar to the ambient conditions of $\vec{v}$. We found that the optimal value of $k_2$ is 6 (see Section 6.2). Let $K_2$ be the set of the $k_2$ baselines that are the next most similar to $\vec{v}$. Note that $K_1$ does not overlap with $K_2$. For each element $v_i$ of $\vec{v}$, the expected variation $\sigma_i(\vec{c})$ is the root mean square of the difference between $\mu_i(\vec{c})$ and the value of the corresponding element in each of the members of $K_2$.

The intuition here is that $K_1$ contains the closest neighbors to $\vec{v}$, so $K_1$ is most suitable for calculating the expected value $\mu_i(\vec{c})$ of $v_i$. $K_2$ contains the next closest neighbors to $\vec{v}$, thus $K_2$ surrounds $K_1$ like a halo, as shown in Figure 3. For illustration purposes, Figure 3 only shows two of the five dimensions of context space. We omit the scale, since Figure 3 does not display the actual locations of the baseline observations.

> Place Figure 3 here.

By comparing $K_2$ with $K_1$, we obtain an estimate of the amount of variation $\sigma_i(\grave{c})$ in the general region of $v_i$. The motivation for making $K_1$ disjoint from $K_2$ is that it increases the estimated variation $\sigma_i(\grave{c})$. We believe that it is safer to overestimate the variation than to underestimate it. Alternatives would be to make $K_1$ a subset of $K_2$ or to make $K_1$ equal to $K_2$. We did not explore these possibilities, since we expected them to be inferior.

### 4.2 Phase 2: Diagnosis with IBL

Suppose we have a normalized feature vector $\vec{\grave{\eta}}$ from the testing set, and we wish to classify $\vec{\grave{\eta}}$ into one of the eight classes. The prediction for $\vec{\grave{\eta}}$ is the class of the majority of the $k_3$ most similar normalized feature vectors in the training set. The measure of similarity is given in Equation 2. We found that the optimal value of $k_3$ is 1 (see Section 6.2). When there is a tie (when there is no majority), we use the degree of similarity to break the tie.

## 5  Linear Regression

We also implemented Phases 1 and 2 using multivariate linear regression [5]. Like IBL, MLR is a natural candidate, since it meets our requirements and it is a popular traditional statistical technique. We call the MLR implementation of our strategy CNMLR.

### 5.1  Phase 1: Feature Normalization with MLR

Suppose we have a feature vector $\vec{\grave{v}}$ that we wish to normalize, using Equation 1. We need to find the expected value $\mu_i(\grave{c})$ and expected variation $\sigma_i(\grave{c})$ of each element $v_i$ of $\vec{\grave{v}}$. We calculate the expected value $\mu_i(\grave{c})$ by using a linear equation. We have one linear equation for each element $v_i$ of $\vec{\grave{v}}$. The independent variables in the equations are a subset of the ambient conditions. The subset is chosen using an automatic variable selection procedure, called *step-wise variable selection* [5]. This procedure has a parameter, which is a threshold $f$ for the $F$-statistic. We found that the optimal value of $f$ is 5 (see Section 6.2). We also allow the linear equations to contain a constant, so an equation may have as many as six terms. The coefficients of the linear equations are set by applying linear regression techniques to the 16 baseline feature vectors.

To calculate the expected variation $\sigma_i(\vec{c})$, we use the formula

$$\sigma_i(\vec{c}) = \sqrt{\sum_{j=1}^{s} \frac{(b_{ij} - \mu_i(\vec{c}_j))^2}{(s-t)}} \qquad \text{(EQ 3)}$$

where $b_{ij}$ is the $i$-th element of the $j$-th baseline vector, $\vec{c}_j$ is the context vector for the $j$-th baseline vector, $s$ is the number of baseline vectors (16), and $t$ is the number of terms in the linear equation for $\mu_i(\vec{c})$ (from 1 to 6). This is a standard technique in linear regression for calculating variation [6]. Note that $\vec{c}$ is not actually used in Equation 3, so the expected variation $\sigma_i(\vec{c})$ is not actually context-sensitive, although the expected value $\mu_i(\vec{c})$ is context-sensitive.

## 5.2 Phase 2: Diagnosis with MLR

In Phase 2, we have one linear equation for each of the eight classes. The independent variables in the equations are a subset of the elements of the normalized feature vectors. The subset is chosen using an automatic variable selection procedure. We used a modified form of *forward variable selection* [5]. Step-wise variable selection was not successful, because two of the eight classes had only two to three observations in the training set. In these cases, none of the variables were judged to be statistically significant, according to the *F*-statistic. Instead of testing for significance, we selected variables until $m$ variables were in the equation. We found that the optimal value of $m$ is 1 (see Section 6.2). All equations contained a constant term, in addition to the $m$ selected terms.

The dependent variable in the equation for class X is set to 1 in the training set when the normalized feature vector belongs to class X, otherwise 0. Suppose we have a normalized feature vector $\vec{\eta}$ from the testing set, and we wish to classify $\vec{\eta}$ into one of the eight classes. Our prediction for $\vec{\eta}$ is the class whose equation's prediction is closest to 1.

## 6 Experimental Results

This section presents the results of a series of experiments that we performed. We were interested in three main questions: How does CNMLR compare with CNIBL? Does contextual normalization (Phase 1) improve the accuracy of diagnosis (Phase 2)? Most importantly, can our approach form the basis of practical software for diagnosis of aircraft gas turbine engines?

## 6.1  Method of Evaluation

If our algorithms are to be the basis for software that will be used by engine repair technicians in the field, then we must be able to recognize a fault in the winter, when we have only seen it previously in the summer, and vice versa. Therefore we do not follow the usual testing methodology of randomly splitting the data into training and testing sets. We split our observations into two sets of roughly equal size. The first set was observed in October and the second set was in November. When our data were collected, it was much warmer in October than it was in November. One set was made the training set and the other was the testing set. The algorithms were evaluated on the testing set, then the two sets were swapped and the algorithms were evaluated again.

We used two scores to evaluate the algorithms. The first score, which we call the *raw* score, was the percent of correct classifications in the combination of the two tests. The raw score is biased, since some classes were much larger than others. An algorithm could get a deceptively high score by doing well with the larger classes and doing poorly with the smaller classes. For example, if 90% of the engines are healthy, then the trivial algorithm that always guesses "healthy" will get a raw score of 90%. We devised a second score, which we call the *adjusted* score, that compensated for the different sizes of the classes. Each class X was scored separately by the average of the probability that the algorithm guessed X, given that the class was actually X, and the probability that the class was actually X, given that the algorithm guessed X. The adjusted score is the average of these individual scores for the eight classes.

## 6.2  Comparison of CNIBL and CNMLR

Table 3 shows the score of CNIBL when using the parameter values $k_1 = 2$, $k_2 = 6$, and $k_3 = 1$. Table 4 shows the score of CNMLR when using the parameter values $f = 5$ and $m = 1$. CNIBL's raw score was 64.0% and its adjusted score was 63.8%. CNMLR's raw score was 51.7% and its adjusted score was 36.7%. For our task, CNIBL is clearly superior to CNMLR.

Place Table 3 here.

> Place Table 4 here.

Tables 3 and 4 show that CNIBL was relatively insensitive to the sizes of the classes, while CNMLR was quite sensitive. CNMLR did as well as CNIBL on the two largest classes, but performed comparatively poorly on the other classes. We expect that the scores of CNIBL and CNMLR would converge with large samples of all of the classes. The advantage of CNIBL is that it does well with small samples (compared to CNMLR).

In timing tests, there was little difference between CNIBL and CNMLR. Our algorithms were written in PV-WAVE™ [7]. Timing tests were done on a Sun Sparc 1. No special effort was made to optimize the efficiency. CNIBL takes less time to train, but more time to test, than CNMLR. Processing a single observation, from Phase 0 to Phase 2, including testing and training, requires an average of 42 seconds (elapsed time, not CPU time) for both CNIBL and CNMLR. Phase 0 takes 40 of those 42 seconds.

We noted in Section 3.1 that the feature detection routines in Phase 0 sometimes do not find a feature and sometimes make mistakes, so the output of Phase 0 includes missing and erroneous values. We experimented with several strategies for handling missing and erroneous values. The most effective strategy was the following. When an engine is unhealthy, the x-axis locations of features tend to increase (that is, the timing of a feature such as "the peak in THRUST" tends to be delayed) and the y-axis locations of features tend to decrease (for example, the force of "the peak in THRUST" tends to be decreased). When a feature is missing, we assume that it is missing because the engine is so unhealthy that the subroutine for detecting the feature could not recognize the feature. Therefore we set the missing feature's normalized $x$ value to $d$ and the normalized $y$ value to $-d$. Figure 4 illustrates this method. (Recall from Section 3.2 that the normalization of Phase 1 is centered on zero.) The value of $d$ is an adjustable parameter for CNIBL and CNMLR. Note that the features for an unhealthy engine do not always move downwards and towards the left. This is merely the most common trend in our data.

Place Figure 4 here.

It is more difficult to handle erroneous values than missing values, since it is obvious when a feature is missing, but it is more difficult to know when a feature is erroneous. We assume that, when a normalized feature value is outside of the range $[-d, d]$, then it is erroneous, and we treat it the same way that we treat a missing value.

Figure 5 shows the performance of CNIBL and CNMLR for various values of the parameter *d*. The best results for CNIBL were obtained with *d* set to 50 (or 55) and the best results for CNMLR were obtained with *d* set to 15. Note that Tables 3 and 4 are based on these optimal settings for *d*. Figure 5 shows that CNIBL consistently performs better than CNMLR, for values of *d* from 5 to 80.

Place Figure 5 here.

It is natural to wonder whether the superior performance of CNIBL is due to its performance on Phase 1 or its performance on Phase 2. One way to address this question is to make a hybrid system. Figure 6 shows the performance of hybrid systems for various values of *d*, from 15 to 50. IBL-1/MLR-2 is a hybrid system using IBL for Phase 1 and MLR for Phase 2. Similarly MLR-1/IBL-2 is a hybrid system using MLR for Phase 1 and IBL for Phase 2.

Place Figure 6 here.

In Figure 6, the best raw score for IBL-1/MLR-2 is 51.2% and the best adjusted score is 26.8%. The best raw score for MLR-1/IBL-2 is 56.2% and the best adjusted score is 41.7%. Comparing the hybrid scores in Figure 6 with the purebred scores in Figure 5, we see that IBL is clearly superior for Phase 2, since the worst scores with IBL for Phase 2 (CNIBL and MLR-1/IBL-2) are better than the best scores with MLR for Phase 2 (CNMLR and IBL-1/MLR-2).

However, there is no clear victor for Phase 1. When IBL is used for Phase 2, IBL is superior to MLR for Phase 1, since the worst score for CNIBL is better than the best score for MLR-1/IBL-2. When MLR is used for Phase 2, it is not clear what is best for Phase 1, since there is some overlap in the scores for CNMLR and IBL-1/MLR-2. We conclude that IBL is superior to MLR for prediction of class membership (Phase 2), but neither is clearly superior for prediction of real numbers (Phase 1).

CNIBL has parameters $k_1$ and $k_2$ in Phase 1 and $k_3$ in Phase 2. CNMLR has $f$ in Phase 1 and $m$ in Phase 2. The values we mentioned in Sections 4 and 5 are the values that gave the highest scores (both raw and adjusted). We investigated the sensitivity of the algorithms to the parameter settings. For Phase 1, we experimented with $k_1 = 1, 2, 3, \ldots, 8$, $k_2 = 4, 6, 8$, and $f = 1, 2, 3, \ldots, 8$. During these experiments, we kept $m$ and $k_3$ fixed at their optimal values ($m = 1$ and $k_3 = 1$). Figure 7 shows the results for CNIBL and Figure 8 shows the results for CNMLR. The best values for CNIBL were $k_1 = 2$ and $k_2 = 6$. The best value for CNMLR was $f = 5$.

Place Figure 7 here.

Place Figure 8 here.

Figures 7 and 8 show that CNIBL was consistently better than CNMLR. There were 24 experiments for CNIBL (8 $k_1$ settings and 3 $k_2$ settings = 24 combined settings). In 22 of these experiments, the adjusted score was greater than the best adjusted score for CNMLR (36.7%).

For Phase 2, we experimented with $k_3 = 1, 2, 3, 4, 5$, and $m = 1, 2, 3, 4, 5$. During these experiments, we kept $k_1$, $k_2$, and $f$ fixed at their optimal values ($k_1 = 2$, $k_2 = 6$ and $f = 5$). Figure 9 shows the results for CNIBL and Figure 10 shows the results for CNMLR. The best value for CNIBL was $k_3 = 1$. The best value for CNMLR was $m = 1$. Note that, in Figure 9, we have the same score for $k_3 = 1$ and $k_3 = 2$. This is because IBL uses majority voting in Phase 2, with ties broken by the degree of similarity. Majority voting with the two most similar neighbors is thus the same as taking the single most similar neighbor.

Place Figure 9 here.

Place Figure 10 here.

Again, Figures 9 and 10 show that CNIBL was consistently better than CNMLR. The lowest adjusted CNIBL score of 47.2% (with $k_3 = 5$) was better than even the highest adjusted CNMLR score of 36.7% (with $m = 1$).

It is natural to wonder why CNIBL performs better than CNMLR on Phase 2. CNIBL performs best on Phase 2 when the single nearest neighbor is used (when $k_3 = 1$). This suggests that the normalized vectors generated in Phase 1 are scattered in the normalized feature space; they are not concentrated in clusters, where the vectors in a cluster all belong to the same class. Scattered vectors are particularly difficult for MLR, since MLR classifies observations by separating them with a hyperplane [5]. Clusters can often be separated by a hyperplane, but a hyperplane will not work well with scattered vectors.

To test this hypothesis, we listed the five nearest neighbors in the training set for each normalized feature vector in the testing set. The classes of the five nearest neighbors tended to be highly heterogenous. This low level of clustering explains why CNIBL performs best on Phase 2 when $k_3 = 1$ and it explains why CNIBL performs significantly better than CNMLR.

In summary, these results show that CNIBL is consistently better than CNMLR, for our problem, for a wide range of the parameters $k_1$, $k_2$, and $k_3$ of CNIBL, $f$ and $m$ of CNMLR, and $d$ of CNIBL and CNMLR. The superiority of CNIBL seems largely due to its performance on Phase 2, compared to the performance of CNMLR on Phase 2. The superiority of CNIBL seems to be due to the ability of IBL to handle data with a low level of clustering. We expect IBL to perform better than MLR in other applications where the data have a low level of clustering. We conjecture that many machinery diagnosis problems share this property (a low level of clustering) with gas turbine engine diagnosis.

## 6.3  Testing Contextual Normalization

It is natural to consider whether contextual normalization is necessary. We experimented with the following methods for normalizing the output of Phase 0:

1. no normalization (that is, take feature vectors straight from Phase 0)

2. normalization by minimum and maximum in the training set

3. normalization by average and standard deviation in the training set

4. normalization by percentile in the training set (e.g. if 10% of the numbers in the training set are below a given value, then that value is normalized to 0.1)

5. normalization by the average and standard deviation in the healthy baseline set

6. normalization by a learning algorithm that uses IBL to predict the value and variation of each feature, for a healthy engine, as a function of the ambient conditions (see Section 4.1)

7. normalization by a learning algorithm that uses MLR to predict the value and variation of each feature, for a healthy engine, as a function of the ambient conditions (see Section 5.1)

Note that only 6 and 7 involve learning and contextual normalization.

The method we developed for dealing with missing and erroneous values in the output of Phase 0 (see Section 6.2) is only suitable for features that have been normalized by methods 5, 6, or 7. Therefore, in order to compare the above methods of normalization, we considered three different ways of handling missing values:

1. set a missing value to zero

2. set a missing value to the average of the value in the training set

3. set a missing $x$ value to the maximum of the value in the training set and set a missing $y$ value to the minimum of the value in the training set

The first method is the simplest, but also the most arbitrary. There is no justification for replacing missing values with zero, except that it is necessary to replace them with *some* value. The second method is intuitively appealing and it is commonly used. The third method uses background knowledge of the behavior of sick engines. It is similar to the method of Section 6.2. To understand the intuition behind the third method, examine Figure 4.

We cannot simply ignore missing values, but we can ignore erroneous values. We decided to handle erroneous values by assuming that there were no erroneous values. That is, we assumed that the feature detectors in Phase 0 would either find the correct feature or report that the feature was missing.

We performed 42 experiments to test the utility of contextual normalization. We tested each of the 7 normalization methods with each of the 3 ways of handling missing values and both of the 2 methods (IBL and MLR) for Phase 2 ($7 \times 3 \times 2 = 42$). Table 5 summarizes the results of the 42 experiments.

Place Table 5 here.

Note that MLR is not affected by linear transformations of the data. Normalization methods 2, 3, and 5 involve a linear transformation. Thus the score for MLR with methods 2, 3, and 5 is the same as the score when there is no treatment of the data (normalization method 1).

Table 5 shows that, when Phase 2 is IBL, the best result (adjusted score 57.4%) is obtained when we use IBL in Phase 1 and handle missing values by method 3. When Phase 2 is MLR, the best result (adjusted score 31.3%) is obtained when we use MLR in Phase 1 and handle missing values by method 3. These results suggest that Phase 1 benefits from the use of a learning strategy. That is, normalization by methods 6 and 7 appears to be superior to the alternatives.

Let us apply a more rigorous statistical analysis to Table 5. We have two hypotheses to test:

1. Normalization by method 6 is superior to normalization by methods 1 to 5.

2. Normalization by method 7 is superior to normalization by methods 1 to 5.

To test hypothesis 1, we compare each row in Table 5 for normalization methods 1 to 5 to the corresponding row in Table 5 for normalization method 6. For example, consider the second row in Table 5. Normalization method 1 is used with IBL in Phase 2 and missing values are handled by method 2. The adjusted score for this row is 22.0%. The corresponding row for normalization method 6 is the second row of the group with normalization method 6. In this row, normalization method 6 is used with IBL in Phase 2 and missing values are handled by method 2. The adjusted score for this row is 53.4%. We are interested in the ratio of these adjusted scores, 53.4% / 22.0% = 242.7%. This ratio indicates that, when we use IBL for Phase 2 and method 2 to handle missing values, normalization method 6 is 142.7% better than normalization method 1. We calculate this ratio for all 30 rows in Table 5 that use normalization methods 1 to 5 ($5 \times 3 \times 2 = 30$). We then apply the Student *t*-test at 95% confidence [6]: There is a 95% probability that normalization method 6 is more than 21.8% better than methods 1 to 5. We use the same technique to evaluate the second hypothesis: There is a 95% probability that method 7 is more than 30.8% better than methods 1 to 5. Hypotheses 1 and 2 are confirmed.

There is no significant difference between the adjusted scores for methods 6 and 7, according to the Student *t*-test. This is what we would expect from Section 6.2, where we saw that IBL and MLR performed comparably for Phase 1. The advantage of IBL over MLR is only significant in Phase 2.

Table 5 seems to indicate that a purebred system (IBL for both phases or MLR for both phases) is superior to a hybrid system (IBL for Phase 1 and MLR for Phase 2, or vice versa). According to the Student *t*-test, there is a 95% probability that purebred systems are more than 33.7% better than hybrid systems. We have no explanation for this observation. The observation is confirmed by comparing Figures 5 and 6.

Table 5 shows that missing values are best handled by method 3. Note that the best adjusted score for method 3 is 57.4% for IBL and 31.3% for MLR. When we handle missing values by the method of Section 6.2, the best adjusted score is 63.8% for IBL and 36.7% for MLR. Thus the method of Section 6.2 adds about 5% to the adjusted scores, when compared with method 3. The method of Section 6.2 is significantly better than the three methods examined in this section, according to the Student *t*-test.

From these results, we conclude that contextual normalization is better than other common forms of normalization, for the given problem. In Section 3.2, we described three characteristics of the learning algorithm in Phase 1. The third characteristic we mentioned is that Phase 1 compensates for engine performance variations that are due to variations in the ambient conditions. None of the normalization methods 1 to 5 have this particular characteristic. This characteristic is vital, given the testing methodology described in Section 6.1. If we used an alternative testing methodology, such as randomly selecting the testing and training samples, the advantage of contextual normalization might not be as apparent. However, if our work is to be applicable to the real world, then our algorithm must be able to (for example) diagnose a fault in the winter when it has only previously seen that fault in the summer. This requirement is what makes contextual normalization necessary.

Many different applications may benefit from contextual normalization, particularly when the context in the testing set differs from the context in the training set. We give some examples of such applications in Section 9.

### 6.4 Practical Evaluation

Our results are not directly comparable with the performance of repair technicians. Currently, engine repair technicians do not go directly from analysis of the data to a diagnosis. The data are visually inspected and symptoms are noted. The symptoms suggest a possible diagnosis, which is tested by gathering sensor information while the engine goes through a prescribed sequence of operations. If the test confirms the possible diagnosis, then the suspect component is replaced. This procedure is iterated until the engine appears to be healthy. Experienced engine repair technicians believe that CNIBL is performing at a level where it can play a useful role in this repair procedure. The output of Phase 1 can assist in the identification of symptoms and the output of Phase 2 can be taken as a possible diagnosis.

In an empirical evaluation of the type reported here, it is important to compare the results with random guessing. For example, suppose we have two classes, A and B. Imagine that we have an algorithm that receives a raw score of 96% when evaluated on a test set. This seems like a good score, but we must compare it with random guessing. Suppose 95% of our observations are in class A and the remaining 5% are in class B. If we adopt the strategy of always guessing that an observation belongs to class A, then we will get a raw score of 95%. In this case, the score of 96%, received by the algorithm, no longer appears impressive.

Table 3 shows that CNIBL's raw score was 64.0% and its adjusted score was 63.8%. To maximize the raw score for random guessing, we should always guess the most common class. In our data, the most common class is the healthy class, class 8. If we always guess class 8, then we will get a raw score of 32.2%. To maximize the adjusted score for random guessing, we should guess each class randomly, with a probability equal to the frequency of the class. The expected adjusted score for this method of random guessing is 12.5%. Thus CNIBL is performing at a rate that is much better than chance.

By examining Tables 3 and 4, we can find the classes that are hardest for CNIBL and CNMLR to discriminate. Ideally, the observations should cluster on the diagonals in the tables (the diagonals have been shaded for emphasis). The off-diagonal values should all be zero. We see from Tables 3 and 4 that the major problem area for CNIBL and CNMLR is the row for class 8. This means that the algorithms tend to classify an observation as belonging to class 8 (the healthy class), when the observation does not belong to class 8.

We have an explanation for this pattern. Classes 6 and 7 represent faults that are binary: They are either present or absent. Classes 1 to 5, on the other hand, represent faults that have various degrees of severity (see Table 1). For example, class 1 is a leak in the P3-MFC line. This leak was deliberately implanted by adding a bleed valve to the P3-MFC line. For different observations of class 1, the bleed valve was opened to different degrees. In some cases, the degree of fault that was implanted was so small that it should not really count as a fault. For example, when the bleed valve in the P3-MFC line is open a tiny amount, the engine technicians reported that it had no effect on the engine. These small "faults" are the major cause of mistaken classification for CNIBL and CNMLR. The "faults" that are missed by CNIBL are essentially the same as those that the technicians reported as having no effect.

To test this, we eliminated less severe faults from the 242 observations. We removed 6 observations from class 1, 10 observations from class 3, 19 observations from class 4, and 3 observations from class 5. A total of 38 observations were removed from the original set of 242. Tables 6 and 7 show the performance of CNIBL and CNMLR on the reduced set of observations. If we compare Tables 6 and 7 to Tables 3 and 4, we see that the adjusted score of CNIBL has improved from 63.8% to 68.9%. The adjusted score of CNMLR has improved from 36.7% to 37.5%. The improvement in the raw scores is larger. The raw score of CNIBL improved from 64.0% to 74.0%. The raw score of CNMLR improved from 51.7% to 60.8%. This improvement is close to the maximum improvement that is possible, given the number of observations that were removed.

Place Table 6 here.

> Place Table 7 here.

We conclude that CNIBL can be a valuable tool to assist technicians with aircraft gas turbine engine diagnosis. CNIBL is performing much better than chance. Our data are not entirely realistic, since some of the faults are so subtle that it could be argued that they are not really faults. With more realistic faults, the performance of CNIBL should improve.

As we mentioned in the introduction, CNIBL can be used in conjunction with a knowledge-based system for gas turbine diagnosis [1]. For example, CNIBL can be used to generate preliminary diagnoses (diagnostic hypotheses), which will be further refined by interaction with the knowledge-based system. In this manner, CNIBL can significantly contribute to the value of the knowledge-based system, by accelerating the initial steps of the diagnosis and repair cycle. The level of accuracy that has been demonstrated in this paper is well-suited to this type of application.

Another possibility is to exploit the output of Phase 1; the normalized feature vectors. The engine repair technician can be presented with plots, similar to Figure 2, in which features are flagged when they have normalized values far from zero. For example, all features with normalized values greater than 5 or less than -5 can be colored red. This will immediately draw the repair technician's attention to anomalous sensor data. Such a facility, by itself, can be a valuable aid to the technicians. Note that it is better to place thresholds (such as "greater than 5 or less than -5") on the normalized feature vectors, rather than the feature vectors before normalization, since normalization reduces or removes the context-sensitivity of the features. This enables tighter thresholds to be used, without triggering false alarms. We have found that there is a high correspondence between the features that an experienced technician finds noteworthy and the features that are normalized to values far from zero.

# 7  Related Work

Our three phase procedure is similar to some of the work done with neural networks. Neural networks for reading text often have three phases. In Phase 0, features are extracted from images of letters. In Phase 1, sets of features are classified as letters. In Phase 2, sets of letters are classified as words. Phase 0 is often done outside the neural net, without involving learning [8, 9]. However, although we have three phases, the details of our three phases are clearly quite different from the details of the three phases that are involved in reading text.

Kibler, Aha, and Albert [2] have also compared instance-based learning to linear regression, using data from computer performance specifications and from automobile price information. They focussed on prediction of real numbers. They found that IBL and MLR achieved similar levels of accuracy, but IBL was easier to use, since it required less expert intervention than MLR. Our results suggest that IBL and MLR achieve similar accuracy for prediction of real numbers, but IBL is superior to MLR for prediction of class membership. This is consistent with Kibler *et al.* [2].

Other researchers have applied machine learning techniques to jet engine diagnosis. For example, Dietz, Kiech, and Ali [10] have applied neural networks to jet and rocket engine fault diagnosis. Malkoff [11] has applied neural networks to fault detection and diagnosis in ship gas turbine propulsion systems. Montgomery [12] has applied machine learning techniques to the diagnosis of aircraft gas turbine engine faults. Our work is distinct from this previous work, in that the previous applications of machine learning techniques do not use anything similar to the contextual normalization that CNIBL and CNMLR use in Phase 1. Therefore, we believe that previous approaches would perform relatively poorly when tested according to the methodology of Section 6.1.

Katz, Gately, and Collins [13] have examined the problem of *robust classification*, applied to the classification of infrared and television images. A classifier system is robust if it can perform classification well, even when the context in the testing set is different from the context in the training set. Katz *et al.* describe three strategies that a classifier system might use in order to perform robust classification. (1) It can use the context to normalize the features. (2) The feature space can be enlarged by treating the context variables as additional features. (3) A switching mechanism can use the context to choose the appropriate classifier system from a set of classifier systems, where each system in the set has been trained in a different context. Katz *et al.* use method (3). Our contextual normalization is essentially method (1).

Methods (2) and (3) require that the training set contains examples of each class in a variety of contexts. Contextual normalization only requires examples of the healthy class in a variety of contexts (see Section 3.2). Contextual normalization is a form of low-level reasoning by analogy. When we use the healthy class to learn how to normalize, we assume that we can extend this normalization to the faulted classes, by analogy to the healthy class.

The testing methodology of Section 6.1 does not provide examples of each class of fault in a variety of contexts. Therefore method (1), contextual normalization, has an advantage over methods (2) and (3), when tested according to our methodology. As we explained above, our testing methodology was forced on us by the requirement that we should be able to diagnose a fault in the winter when we have only previously seen the fault in the summer (and vice versa). Method (1) is best suited to meeting this requirement.

Mathematical models of gas turbine engines, using thermodynamical principles, are increasingly used in industry. These models are capable of simulating a wide range of typical engine faults, including all of the faults that we have studied (see Table 1). It should be noted that mathematical models do not compete with our empirical approach to diagnosis; the two are complementary. A mathematical model can be thought of as a function that maps engine parameter values to simulated sensor readings. A particular engine fault can be represented by a certain set of parameter values. The problem with mathematical models is that there is no simple method to invert the function, to map sensor readings to parameter values. The solution is to search through parameter space, to find values that can simulate the given sensor readings. One way to perform this search is to simulate a wide variety of faults with the model, storing the simulated sensor data in a library. An empirical algorithm, such as CNIBL, can then be trained using this library.

## 8  Future Work

We will be trying our strategy with techniques other than MLR and IBL. We plan to try neural networks [10, 11], decision tree induction [14], and genetic algorithms [15]. These are popular learning algorithms that seem applicable to our problem.

The faults we implanted in the engine were chosen to be challenging, but the main criterion for choosing faults was that they should be easy to implant and easy to correct. We plan to experiment with more realistic faults, representative of the faults that engine repair technicians encounter in the field.

The feature extraction algorithm in Phase 0 could be more sophisticated. We would like to include a learning component.

All our data were collected from a single engine. We have begun to gather data from a second engine, with the same implanted faults. Different engines, even of the same model, can have significantly different behaviors, due to such factors as the time since the last overhaul or the severity of usage of a particular engine. We want to establish whether faults observed on one engine of a given model can be used to diagnose a second engine of the same model.

We can achieve higher accuracy by combining several observations of the same fault. For example, given three observations of a single fault, we can apply our algorithms and generate three predicted classes. Suppose two of the predictions are class 8 and one is class 7. We can choose between the competing hypotheses by considering their conditional probabilities. The probability of the hypothesis that the fault is actually class 7 is the product:

$$p(\text{predicted} = 8 | \text{actual} = 7) \cdot p(\text{predicted} = 8 | \text{actual} = 7) \cdot p(\text{predicted} = 7 | \text{actual} = 7) \quad \text{(EQ 4)}$$

These conditional probabilities can be calculated easily from the information in Tables 3 and 4.

To test this method of combining observations, we must use data outside of the set of 242 observations that were used to make Tables 3 and 4. We used a small sample of 14 observations of 3 faults. On this small sample, we used CNIBL to make 14 predictions. We combined these predictions using conditional probabilities, calculated from Table 3. This approach appears very promising. We will report further on this when we have more data.

## 9  Conclusion

The core idea of contextual normalization is to normalize the data in a manner that is sensitive to the context in which the data were collected. Contextual normalization is widely applicable to diagnosis and classification problems. It is particularly useful when the context in the testing set is different from the context in the training set. For example, contextual normalization lets us recognize the similarity between a fault in a gas turbine engine observed in the summer and the same fault observed in the winter. As another example, consider data from an electrocardiogram. We can extract key features from the plots and normalize them using healthy baselines. The context (the "ambient conditions") might be the patient's age, weight, sex, height, occupation, and physical fitness. Features that were unusual, given the context, would be normalized to values that are relatively far from zero. We can then diagnose the condition of the patient's heart, by examining the contextually normalized feature vectors. Other examples of problems that can be addressed with contextual normalization are: the diagnosis of spinal problems, given that spinal tests are sensitive to the age of the patient; the recognition of speech, given that different speakers have different voices; and the classification of images, given varying lighting conditions.

We have three main results. First (Section 6.2), IBL is superior to MLR for aircraft gas turbine engine diagnosis, when using the approach described here. We believe that IBL is superior to MLR whenever the data have a low level of clustering. We conjecture that this is common with machinery diagnosis applications. Second (Section 6.3), contextual normalization is superior to other common forms of normalization, for both IBL and MLR, when using the approach described here. We believe that contextual normalization is widely applicable to diagnosis and classification problems. Third (Section 6.4), our algorithms can be the basis for a tool that can be very useful to engine technicians. CNIBL can be used in a stand-alone software package; it can be used to enhance a knowledge-based diagnosis system; or the output of Phase 1 can be used to flag anomalous sensor readings.

## Acknowledgments

The engine data and engine expertise were provided by the Engine Laboratory, Institute for Mechanical Engineering, National Research Council Canada. The Engine Laboratory received funding from the Department of National Defence. We wish to thank Dr. David Aha of the John Hopkins Applied Physics Laboratory, Tim Taylor of Phalanx Research Systems, and our colleagues at the NRC, Jeff Bird, Dr. Sieu Phan, Bob Orchard, Dr. Xueming Huang, Suhayya Abu-Hakima, Rob Wylie, and Jack Brahan, for their helpful comments on earlier versions of this paper. We would also like to thank two anonymous referees of the *Journal of Applied Intelligence* for their helpful comments.

# Appendix: List of Symbols

| | |
|---|---|
| $b_{ij}$ | is the $i$-th element of the $j$-th baseline vector. |
| $\vec{c}$ | is a context vector. |
| $\vec{c}_j$ | is the context vector for the $j$-th baseline vector. |
| $d$ | is the default value, used by CNIBL and CNMLR to handle missing and erroneous features. |
| $\vec{\eta} = (\eta_1, ..., \eta_n)$ | is a vector of normalized features, output from the contextual normalization of Phase 1. |
| $f$ | is a threshold for the $F$-statistic. It is a parameter of CNMLR in Phase 1. |
| $k_1$ | is the size of $K_1$. It is a parameter of CNIBL in Phase 1. |
| $K_1$ | is the set of the $k_1$ baselines that are most similar to $\vec{v}$. |
| $k_2$ | is the size of $K_2$. It is a parameter of CNIBL in Phase 1. |
| $K_2$ | is the set of the $k_2$ baselines that are the next most similar to $\vec{v}$. |
| $k_3$ | is the size of the set of the normalized feature vectors in the training set that are most similar to a given normalized feature vector in the testing set. It is a parameter of CNIBL in Phase 2. |
| $m$ | is the number of terms in the linear equations for CNMLR in Phase 2. All equations contained a constant term, in addition to the $m$ selected terms. |
| $\mu_i(\vec{c})$ | is the expected value of $v_i$ as a function of the context $\vec{c}$. |
| $s$ | is the number of baseline vectors (16). |
| $\sigma_i(\vec{c})$ | is the expected variation of $v_i$ as a function of the context $\vec{c}$. |
| $t$ | is the number of terms in the linear equation for $\mu_i(\vec{c})$ (from 1 to 6). |
| $\vec{v} = (v_1, ..., v_n)$ | is a vector of features, output from the feature extraction algorithm of Phase 0. $v_i$ is an element of $\vec{v}$. |

# List of Figures and Tables

# Figures

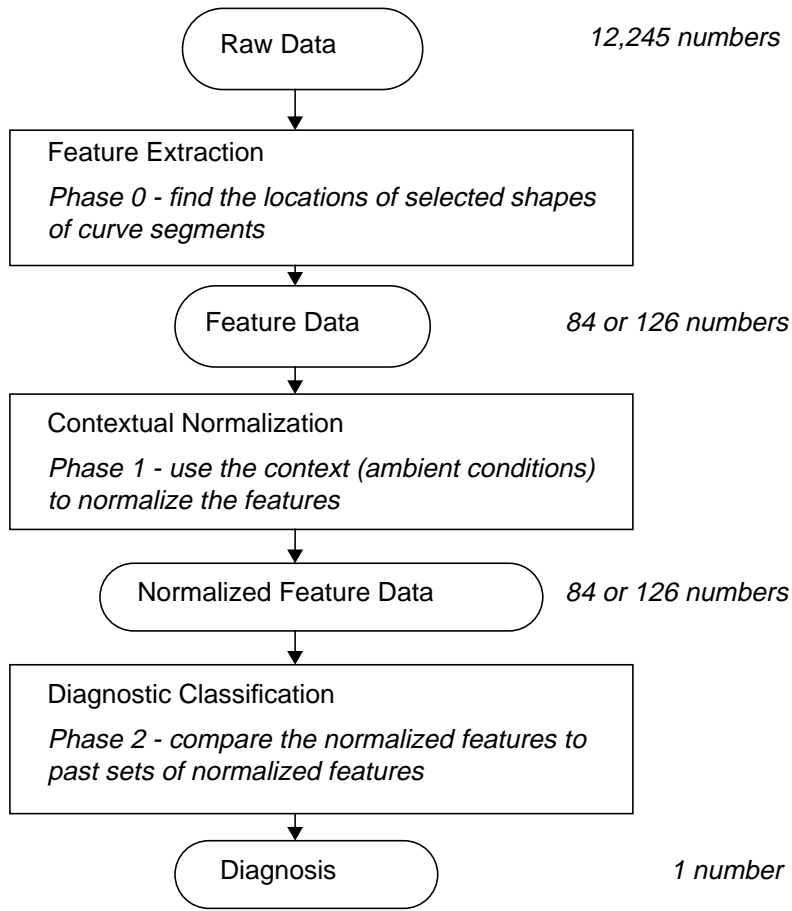

Figure 1. The strategy for data analysis.

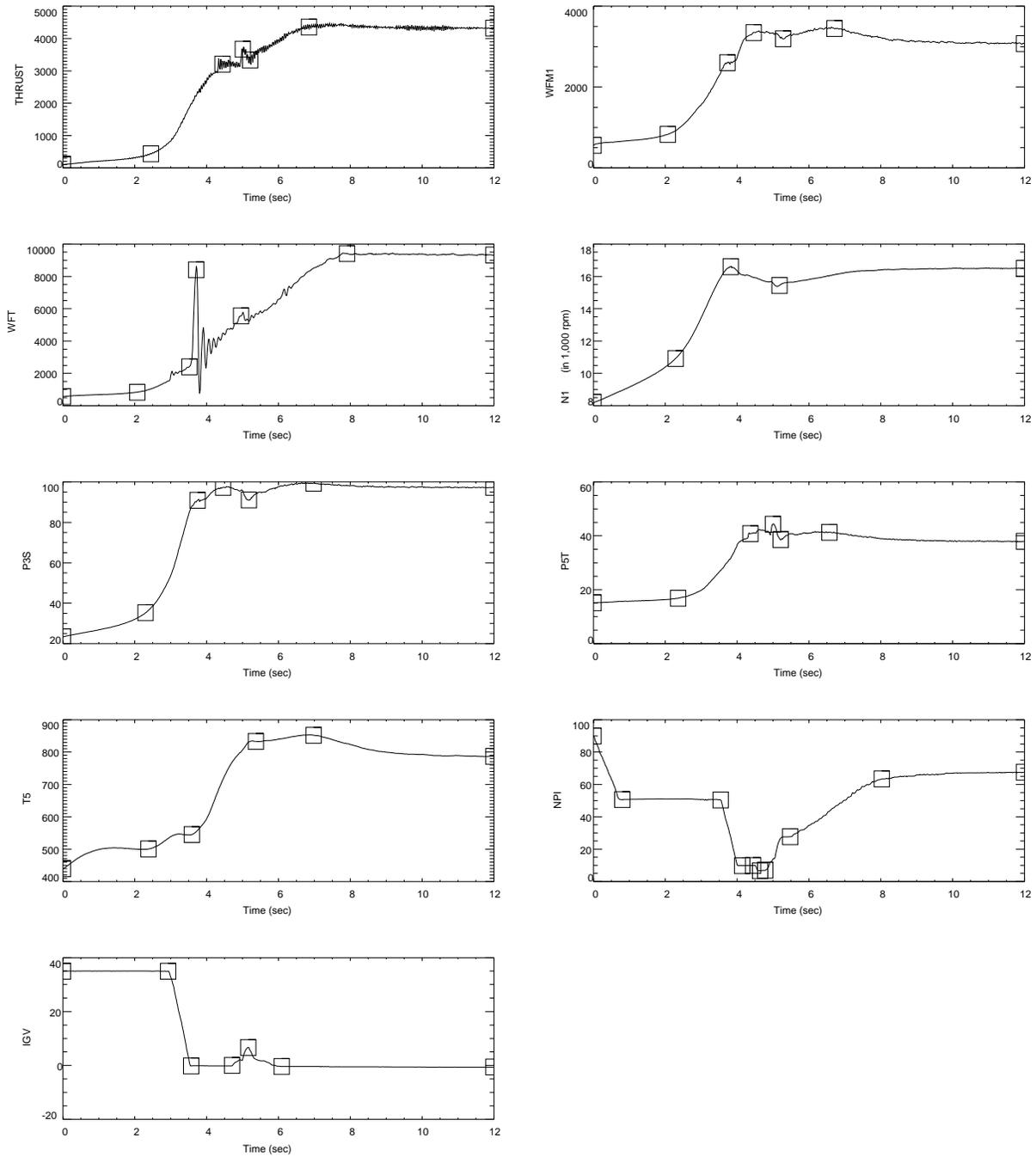

Figure 2. Features for a healthy idle to maximum afterburner acceleration.

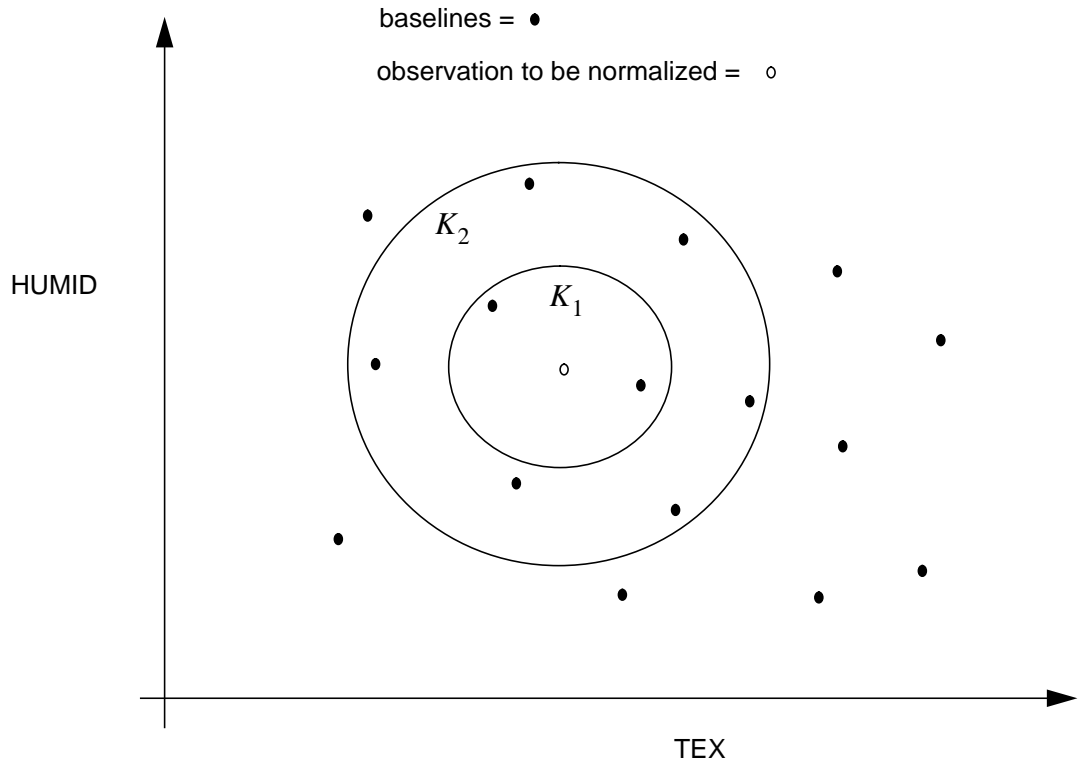

Figure 3. An illustration of $K_1$ and $K_2$ in context space.

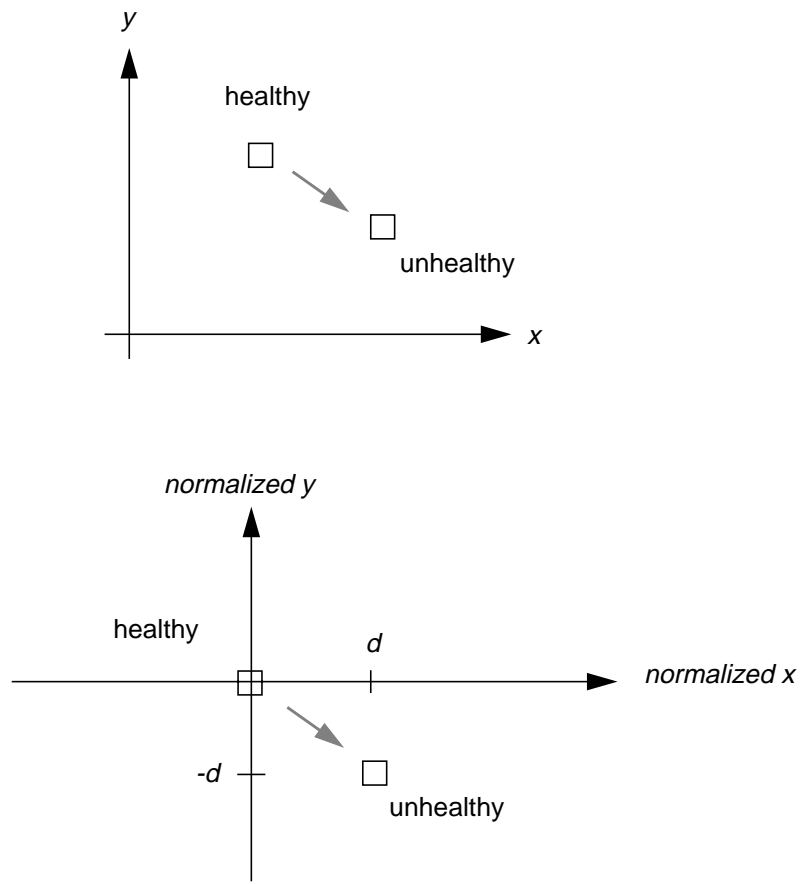

Figure 4. A method for handling missing features.

Figure 5. The performance of CNIBL and CNMLR for various values of *d*.

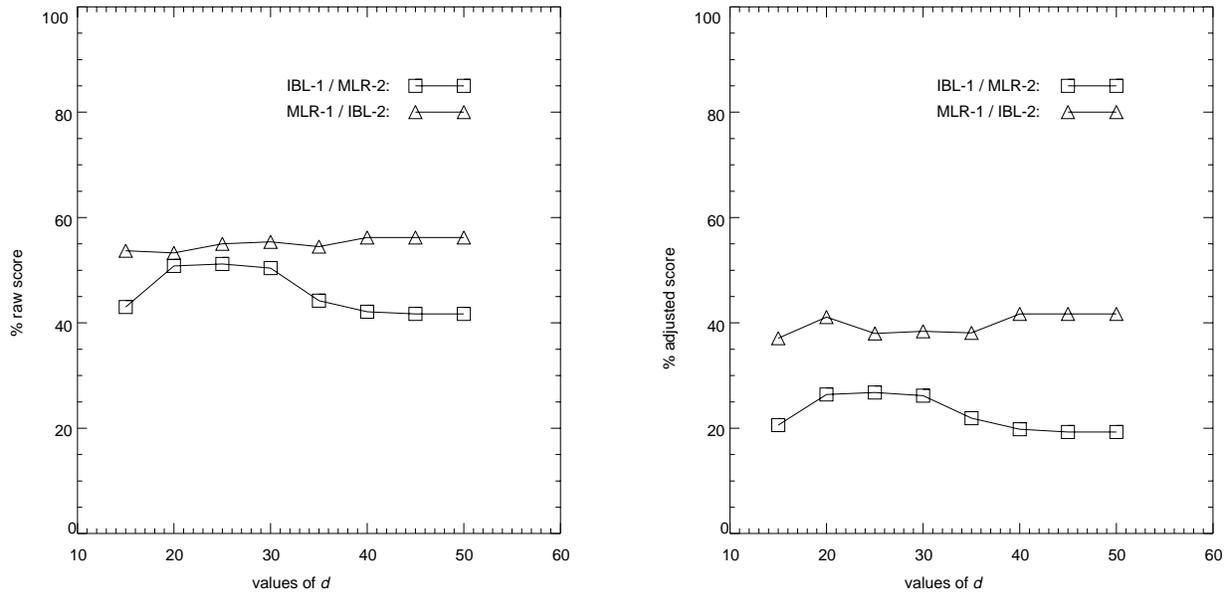

Figure 6. The performance of hybrid systems for various values of *d*.

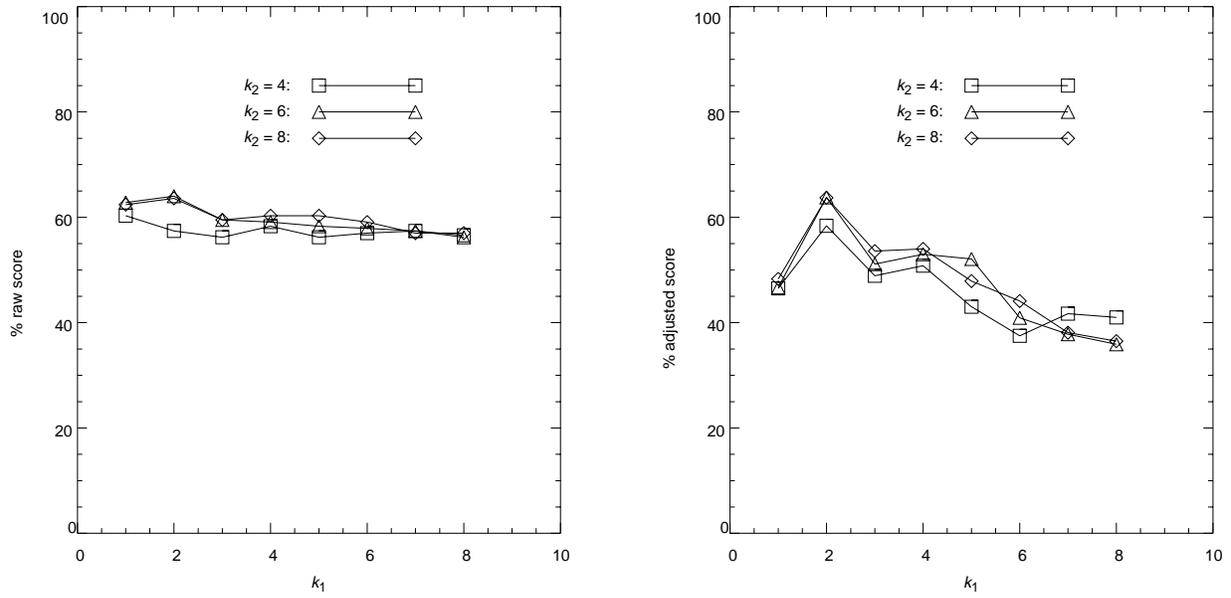

Figure 7. The performance of CNIBL for various settings of $k_1$ and $k_2$.

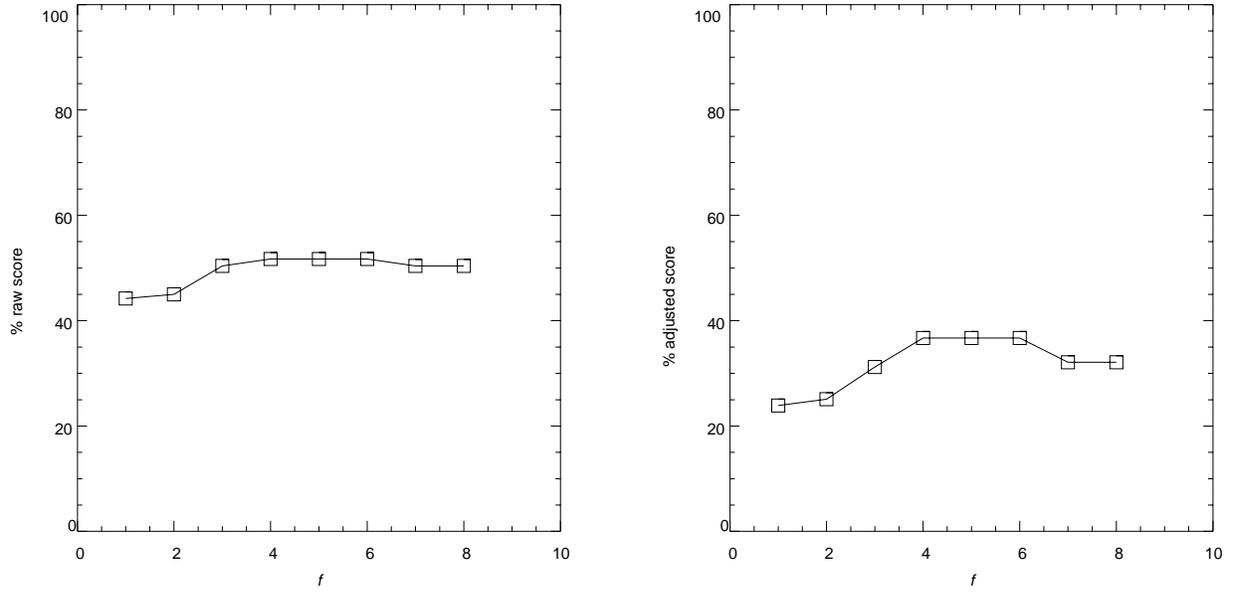

Figure 8. The performance of CNMLR for various settings of $f$.

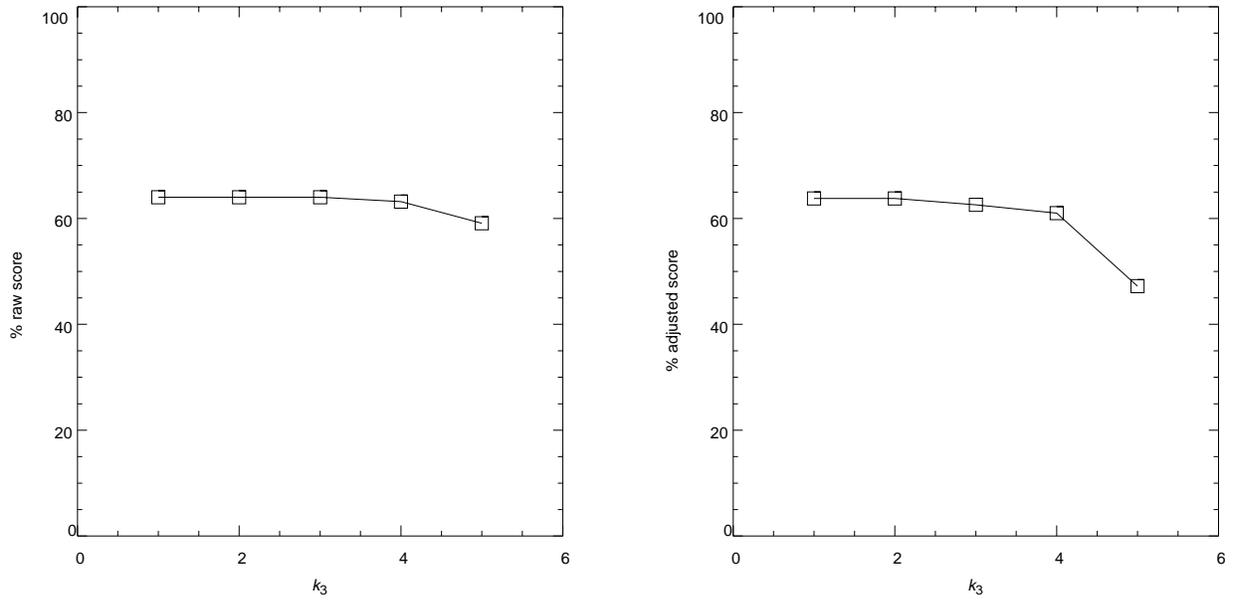

Figure 9. The performance of CNIBL for various settings of $k_3$.

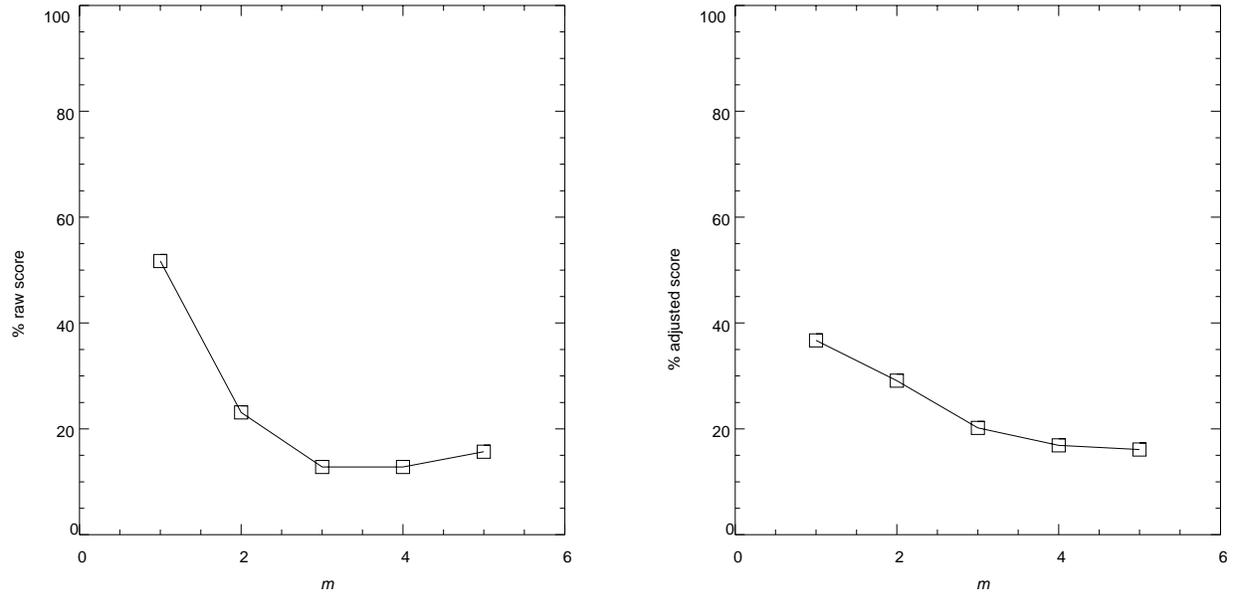

Figure 10. The performance of CNMLR for various settings of $m$.

# Tables

Table 1. The eight classes of observations.

| Class | Description |
|-------|-------------|
| 1 | leak in P3-MFC line |
| 2 | T5 amplifier misadjustment |
| 3 | T5 motor misadjustment |
| 4 | VEN control compensator misadjustment |
| 5 | misadjustment of SG setting of MFC |
| 6 | anti-icing valve left on |
| 7 | afterburner igniter left off |
| 8 | healthy |

Table 2. The variables that were recorded for each observation.

| #  | Symbol | Description               | Units                    | Used |
|----|--------|---------------------------|--------------------------|------|
| 1  | TIME   | time                      | seconds                  | *    |
| 2  | PLA    | power lever angle         | degrees (angular)        |      |
| 3  | N1     | shaft speed               | revolutions per minute   | *    |
| 4  | WFM1   | fuel flow - main          | pounds mass per hour     | *    |
| 5  | WFT    | fuel flow - total         | pounds mass per hour     | *    |
| 6  | T5     | exhaust gas temperature   | degrees Celsius          | *    |
| 7  | IGV    | inlet guide vane position | degrees (angular)        | *    |
| 8  | NPI    | nozzle position indicator | percent                  | *    |
| 9  | THRUST | thrust                    | pounds force             | *    |
| 10 | PBS    | bellmouth static pressure | pounds per square inch   |      |
| 11 | P3S    | compressor delivery pressure | pounds per square inch | *  |
| 12 | WA1R   | airflow                   | pounds mass per second   |      |
| 13 | P5T    | exhaust pressure          | pounds per square inch   | *    |
| 14 | CIT    | compressor inlet temperature | degrees Celsius       |      |
| 15 | DATPT  | data point number         | none                     |      |
| 16 | TFM1   | inlet fuel temperature    | degrees Celsius          |      |
| 17 | T3     | compressor delivery temp  | degrees Celsius          |      |
| 18 | T1     | inlet air temp (constant) | degrees Celsius          | **   |
| 19 | TEX    | outside air temp (constant) | degrees Celsius        | **   |
| 20 | TDEW   | dew point temp (constant) | degrees Celsius          | **   |
| 21 | BARO   | outside air pressure (constant) | pounds per square inch | ** |
| 22 | HUMID  | relative humidity (constant) | percent               | **   |

Table 3. The score of CNIBL with parameter values $k_1 = 2$, $k_2 = 6$, and $k_3 = 1$.

|  |  | \multicolumn{8}{c}{Actual Class of Observation} |  |  |
|---|---|---|---|---|---|---|---|---|---|---|---|
|  |  | class 1 | class 2 | class 3 | class 4 | class 5 | class 6 | class 7 | class 8 | total | P2% |
| Predicted Class of Observation | class 1 | 40 | 0 | 1 | 0 | 4 | 0 | 0 | 1 | 46 | 87.0 |
|  | class 2 | 0 | 6 | 0 | 0 | 0 | 0 | 0 | 1 | 7 | 85.7 |
|  | class 3 | 0 | 0 | 12 | 1 | 0 | 0 | 0 | 0 | 13 | 92.3 |
|  | class 4 | 3 | 3 | 5 | 17 | 2 | 0 | 0 | 2 | 32 | 53.1 |
|  | class 5 | 3 | 0 | 3 | 2 | 2 | 0 | 0 | 2 | 12 | 16.7 |
|  | class 6 | 0 | 0 | 0 | 0 | 0 | 2 | 0 | 0 | 2 | 100.0 |
|  | class 7 | 0 | 0 | 0 | 0 | 0 | 0 | 4 | 0 | 4 | 100.0 |
|  | class 8 | 6 | 3 | 15 | 19 | 7 | 3 | 1 | 72 | 126 | 57.1 |
|  | total | 52 | 12 | 36 | 39 | 15 | 5 | 5 | 78 | 242 | 74.0 |
|  | P1% | 76.9 | 50.0 | 33.3 | 43.6 | 13.3 | 40.0 | 80.0 | 92.3 | 53.7 | 63.8 |

P1: $p(\text{prediction} = x \mid \text{actual} = x)$, P2: $p(\text{actual} = x \mid \text{prediction} = x)$

Raw Score: 64.0%, Adjusted Score: 63.8%

Table 4. The score of CNMLR with parameter values $f = 5$ and $m = 1$.

Actual Class of Observation

|  | class 1 | class 2 | class 3 | class 4 | class 5 | class 6 | class 7 | class 8 | total | P2% |
|---|---|---|---|---|---|---|---|---|---|---|
| class 1 | 42 | 0 | 3 | 2 | 0 | 0 | 0 | 1 | 48 | 87.5 |
| class 2 | 0 | 2 | 0 | 0 | 0 | 0 | 0 | 0 | 2 | 100.0 |
| class 3 | 0 | 0 | 3 | 11 | 7 | 0 | 0 | 5 | 26 | 11.5 |
| class 4 | 0 | 1 | 2 | 7 | 0 | 0 | 2 | 3 | 15 | 46.7 |
| class 5 | 0 | 0 | 4 | 0 | 0 | 0 | 0 | 0 | 4 | 0.0 |
| class 6 | 0 | 0 | 0 | 0 | 0 | 0 | 0 | 0 | 0 | 0.0 |
| class 7 | 0 | 0 | 3 | 0 | 0 | 0 | 2 | 0 | 5 | 40.0 |
| class 8 | 10 | 9 | 21 | 19 | 8 | 5 | 1 | 69 | 142 | 48.6 |
| total | 52 | 12 | 36 | 39 | 15 | 5 | 5 | 78 | 242 | 41.8 |
| P1% | 80.8 | 16.7 | 8.3 | 17.9 | 0.0 | 0.0 | 40.0 | 88.5 | 31.5 | 36.7 |

Predicted Class of Observation (row axis label)

P1: $p(\text{prediction} = x \mid \text{actual} = x)$, P2: $p(\text{actual} = x \mid \text{prediction} = x)$

Raw Score: 51.7%, Adjusted Score: 36.7%

Table 5. A comparison of various methods of normalization.

| Normalize | Phase 2 | Missing | Raw Score | Adjusted Score |
|---|---|---|---|---|
| 1. no normalize | 1. IBL | 1. zero | 41.3 | 24.1 |
| 1. no normalize | 1. IBL | 2. average | 40.5 | 22.0 |
| 1. no normalize | 1. IBL | 3. min/max | 42.1 | 28.3 |
| 1. no normalize | 2. MLR | 1. zero | 41.3 | 19.9 |
| 1. no normalize | 2. MLR | 2. average | 36.4 | 18.3 |
| 1. no normalize | 2. MLR | 3. min/max | 41.3 | 20.5 |
| 2. min/max train | 1. IBL | 1. zero | 40.1 | 22.7 |
| 2. min/max train | 1. IBL | 2. average | 37.6 | 29.0 |
| 2. min/max train | 1. IBL | 3. min/max | 41.7 | 33.9 |
| 2. min/max train | 2. MLR | 1. zero | 41.3 | 19.9 |
| 2. min/max train | 2. MLR | 2. average | 36.4 | 18.3 |
| 2. min/max train | 2. MLR | 3. min/max | 41.3 | 20.5 |
| 3. avg/dev train | 1. IBL | 1. zero | 39.7 | 19.5 |
| 3. avg/dev train | 1. IBL | 2. average | 38.4 | 23.4 |
| 3. avg/dev train | 1. IBL | 3. min/max | 40.1 | 24.4 |
| 3. avg/dev train | 2. MLR | 1. zero | 41.3 | 19.9 |
| 3. avg/dev train | 2. MLR | 2. average | 36.4 | 18.3 |
| 3. avg/dev train | 2. MLR | 3. min/max | 41.3 | 20.5 |
| 4. percent train | 1. IBL | 1. zero | 38.4 | 25.4 |
| 4. percent train | 1. IBL | 2. average | 38.0 | 30.0 |
| 4. percent train | 1. IBL | 3. min/max | 38.0 | 27.8 |
| 4. percent train | 2. MLR | 1. zero | 40.9 | 19.6 |
| 4. percent train | 2. MLR | 2. average | 25.2 | 14.4 |
| 4. percent train | 2. MLR | 3. min/max | 30.6 | 17.0 |
| 5. avg/dev base | 1. IBL | 1. zero | 44.2 | 33.7 |
| 5. avg/dev base | 1. IBL | 2. average | 44.6 | 36.6 |
| 5. avg/dev base | 1. IBL | 3. min/max | 45.9 | 37.6 |
| 5. avg/dev base | 2. MLR | 1. zero | 41.3 | 19.9 |
| 5. avg/dev base | 2. MLR | 2. average | 36.4 | 18.3 |
| 5. avg/dev base | 2. MLR | 3. min/max | 41.3 | 20.5 |
| 6. IBL | 1. IBL | 1. zero | 56.6 | 54.7 |
| 6. IBL | 1. IBL | 2. average | 55.8 | 53.4 |
| 6. IBL | 1. IBL | 3. min/max | 57.4 | 57.4 |
| 6. IBL | 2. MLR | 1. zero | 31.8 | 13.5 |
| 6. IBL | 2. MLR | 2. average | 30.6 | 13.1 |
| 6. IBL | 2. MLR | 3. min/max | 42.6 | 20.0 |
| 7. MLR | 1. IBL | 1. zero | 52.5 | 41.2 |
| 7. MLR | 1. IBL | 2. average | 51.2 | 39.0 |
| 7. MLR | 1. IBL | 3. min/max | 52.9 | 44.4 |
| 7. MLR | 2. MLR | 1. zero | 40.1 | 20.3 |
| 7. MLR | 2. MLR | 2. average | 40.9 | 20.6 |
| 7. MLR | 2. MLR | 3. min/max | 49.2 | 31.3 |

Table 6. The score of CNIBL with less severe faults eliminated.

Actual Class of Observation

|  | class 1 | class 2 | class 3 | class 4 | class 5 | class 6 | class 7 | class 8 | total | P2% |
|---|---|---|---|---|---|---|---|---|---|---|
| class 1 | 39 | 0 | 0 | 0 | 2 | 0 | 0 | 1 | 42 | 92.9 |
| class 2 | 0 | 6 | 0 | 0 | 0 | 0 | 0 | 1 | 7 | 85.7 |
| class 3 | 0 | 0 | 11 | 0 | 0 | 0 | 0 | 0 | 11 | 100.0 |
| class 4 | 3 | 3 | 4 | 14 | 1 | 0 | 0 | 1 | 26 | 53.8 |
| class 5 | 3 | 0 | 2 | 0 | 2 | 0 | 0 | 2 | 9 | 22.2 |
| class 6 | 0 | 0 | 0 | 0 | 0 | 2 | 0 | 0 | 2 | 100.0 |
| class 7 | 0 | 0 | 0 | 0 | 0 | 0 | 4 | 0 | 4 | 100.0 |
| class 8 | 1 | 3 | 9 | 6 | 7 | 3 | 1 | 73 | 103 | 70.9 |
| total | 46 | 12 | 26 | 20 | 12 | 5 | 5 | 78 | 204 | 78.2 |
| P1% | 84.8 | 50.0 | 42.3 | 70.0 | 16.7 | 40.0 | 80.0 | 93.6 | 59.7 | 68.9 |

(Predicted Class of Observation)

P1: $p(\text{prediction} = x \mid \text{actual} = x)$, P2: $p(\text{actual} = x \mid \text{prediction} = x)$

Raw Score: 74.0%, Adjusted Score: 68.9%

Table 7. The score of CNMLR with less severe faults eliminated.

Actual Class of Observation

|  | class 1 | class 2 | class 3 | class 4 | class 5 | class 6 | class 7 | class 8 | total | P2% |
|---|---|---|---|---|---|---|---|---|---|---|
| class 1 | 43 | 0 | 1 | 0 | 3 | 0 | 0 | 1 | 48 | 89.6 |
| class 2 | 0 | 2 | 0 | 0 | 0 | 0 | 0 | 0 | 2 | 100.0 |
| class 3 | 0 | 0 | 3 | 9 | 0 | 0 | 0 | 3 | 15 | 20.0 |
| class 4 | 0 | 1 | 0 | 2 | 0 | 0 | 2 | 2 | 7 | 28.6 |
| class 5 | 0 | 0 | 2 | 0 | 0 | 0 | 0 | 0 | 2 | 0.0 |
| class 6 | 0 | 0 | 0 | 0 | 0 | 0 | 0 | 0 | 0 | 0.0 |
| class 7 | 0 | 0 | 3 | 0 | 0 | 0 | 2 | 0 | 5 | 40.0 |
| class 8 | 3 | 9 | 17 | 9 | 9 | 5 | 1 | 72 | 125 | 57.6 |
| total | 46 | 12 | 26 | 20 | 12 | 5 | 5 | 78 | 204 | 42.0 |
| P1% | 93.5 | 16.7 | 11.5 | 10.0 | 0.0 | 0.0 | 40.0 | 92.3 | 33.0 | 37.5 |

(Predicted Class of Observation — row labels)

P1: $p(\text{prediction} = x \mid \text{actual} = x)$, P2: $p(\text{actual} = x \mid \text{prediction} = x)$

Raw Score: 60.8%, Adjusted Score: 37.5%